\documentclass{article}

\PassOptionsToPackage{numbers}{natbib}
\usepackage[preprint]{tmlr}

\usepackage[utf8]{inputenc} 
\usepackage[T1]{fontenc}    
\usepackage{hyperref}       
\usepackage{url}            
\usepackage{booktabs}       
\usepackage{amsfonts}       
\usepackage{multirow}       
\usepackage{amsmath}        
\usepackage{nicefrac}       
\usepackage{microtype}      
\usepackage{xcolor}         
\usepackage{graphicx}
\usepackage{svg}
\svgsetup{inkscapelatex=false}
\usepackage{wrapfig}
\usepackage{booktabs}
\usepackage{amssymb}  
\usepackage{float}

\title{Geometry-Aware Reinforcement Learning for 2D Irregular Nesting}

\author{\name Auguste Lehuger \email auguste.lehuger@gmail.com \\
      \addr Valeo Brain \\
      \AND
      \name Guillaume Henon-Just \email guillaume.henonjust@gmail.com \\
      \addr Valeo Brain \\}

\begin{document}

\maketitle

\begin{abstract}
Traditional heuristic solvers for the 2D irregular nesting problem share a fundamental limitation: they are blind to polygon geometry, relying on guided brute-force to navigate the continuous placement space with minimal geometrical guidance. In this paper, we argue that Reinforcement Learning is uniquely positioned to overcome this bottleneck. By pairing an optimization policy with a geometry-aware neural encoder, an agent can automatically discover rich geometric priors directly from data, utilizing these learned intuitions to strategically guide exploration. To realize this, we introduce the Polygons Transformer (PoT), a novel architecture that encodes 2D continuous vector geometries while allowing cross-polygon attention. We couple this novel architecture with a Combinatorial Optimization Reinforcement Learning (CORL) training framework to find optimal solutions.
To support this paradigm, we release an open-source training dataset derived from complex geographic contours alongside a dedicated evaluation benchmark. Empirically, our agent slightly exceeds Sparrow, the state-of-the-art heuristic, on small (4-polygon) instances, while a clear scaling gap remains on larger (8-polygon) instances.
\end{abstract}

\section{Introduction}

Combinatorial Optimization problems are traditionally addressed using exact mathematical solvers or handcrafted heuristics, both of which require domain expert knowledge and therefore extensive re-engineering whenever problem constraints shift. To overcome these limitations, Combinatorial Optimization Reinforcement Learning (CORL) has emerged as a promising alternative, leveraging machine learning to automatically discover adaptable search strategies directly from training data \citep{corl}. On canonical graph-based problems such as the Traveling Salesperson Problem or Capacitated Vehicle Routing Problem, CORL approaches now perform competitively, approaching the efficacy of exact solvers with significantly lower inference costs \citep{polynet}. However, to truly validate the relevance of this paradigm, it is crucial to demonstrate its efficiency on problems closer to real-world industrial applications.

In this work, we investigate the application of CORL to a challenging continuous spatial environment: the unconstrained 2D irregular nesting problem. Motivated by Printed Circuit Board (PCB) manufacturing, this problem requires optimizing the area utilization rate of a minimal bounding box enclosing a set of irregular, non-convex polygons. Unlike standard bin packing, this formulation allows for unconstrained boundaries and free continuous spatial rotation and translation of irregular shapes. While our specific industrial applications involve small instances, the multiplier effect of mass production means that improving area utilization by fractions of a percent yields great economic and ecological savings. This incentivizes the allocation of a substantial computational budget per instance to pursue maximum placement precision.

\begin{figure}[htbp]
    \centering
    \begin{tabular}{@{}c@{\hspace{1.2em}}c@{}}
       \raisebox{-\height}{\includegraphics[height=0.25\textwidth]{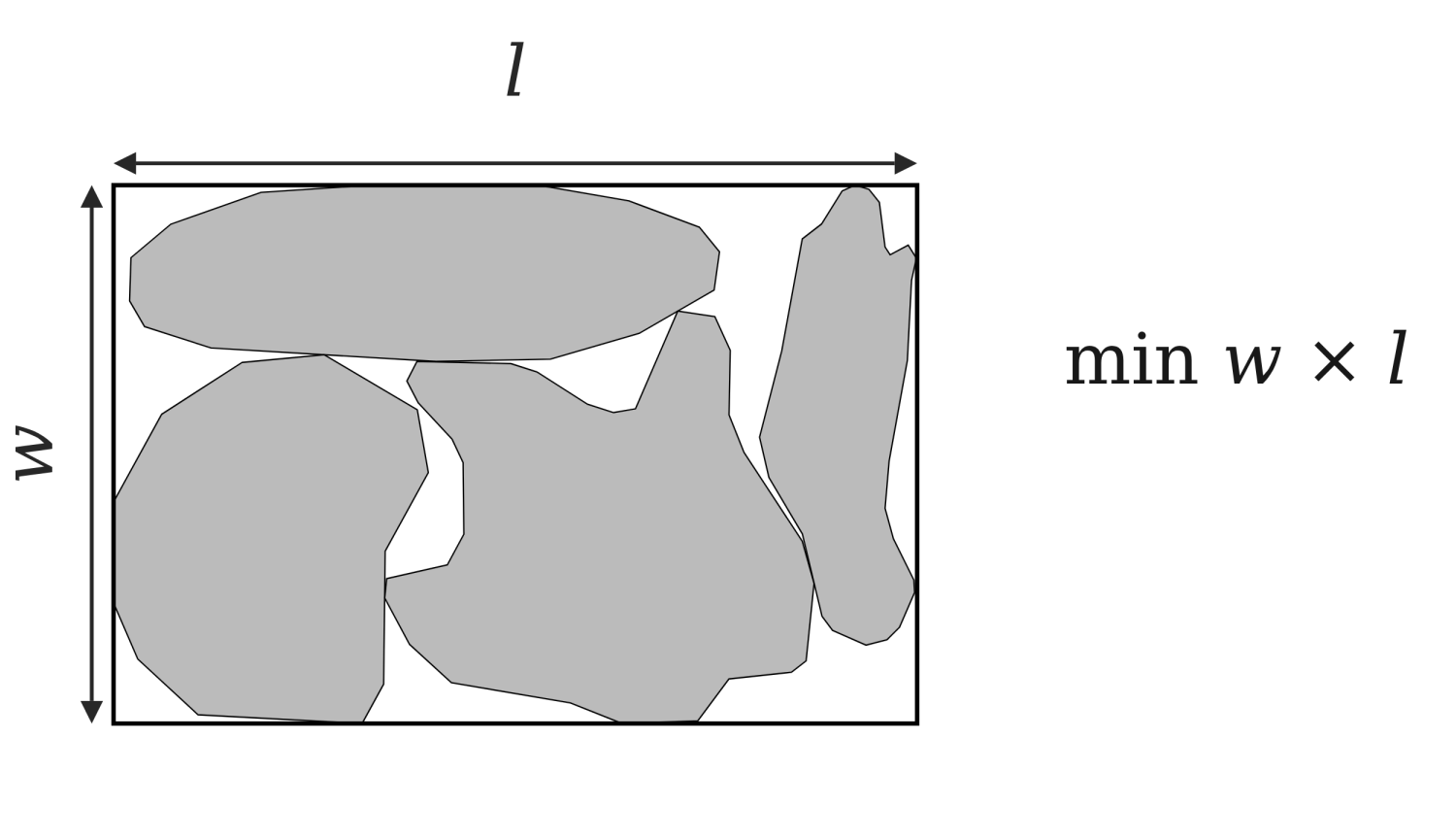}} &
        \raisebox{\dimexpr-\height-0.045\textwidth}{\includegraphics[height=0.185\textwidth]{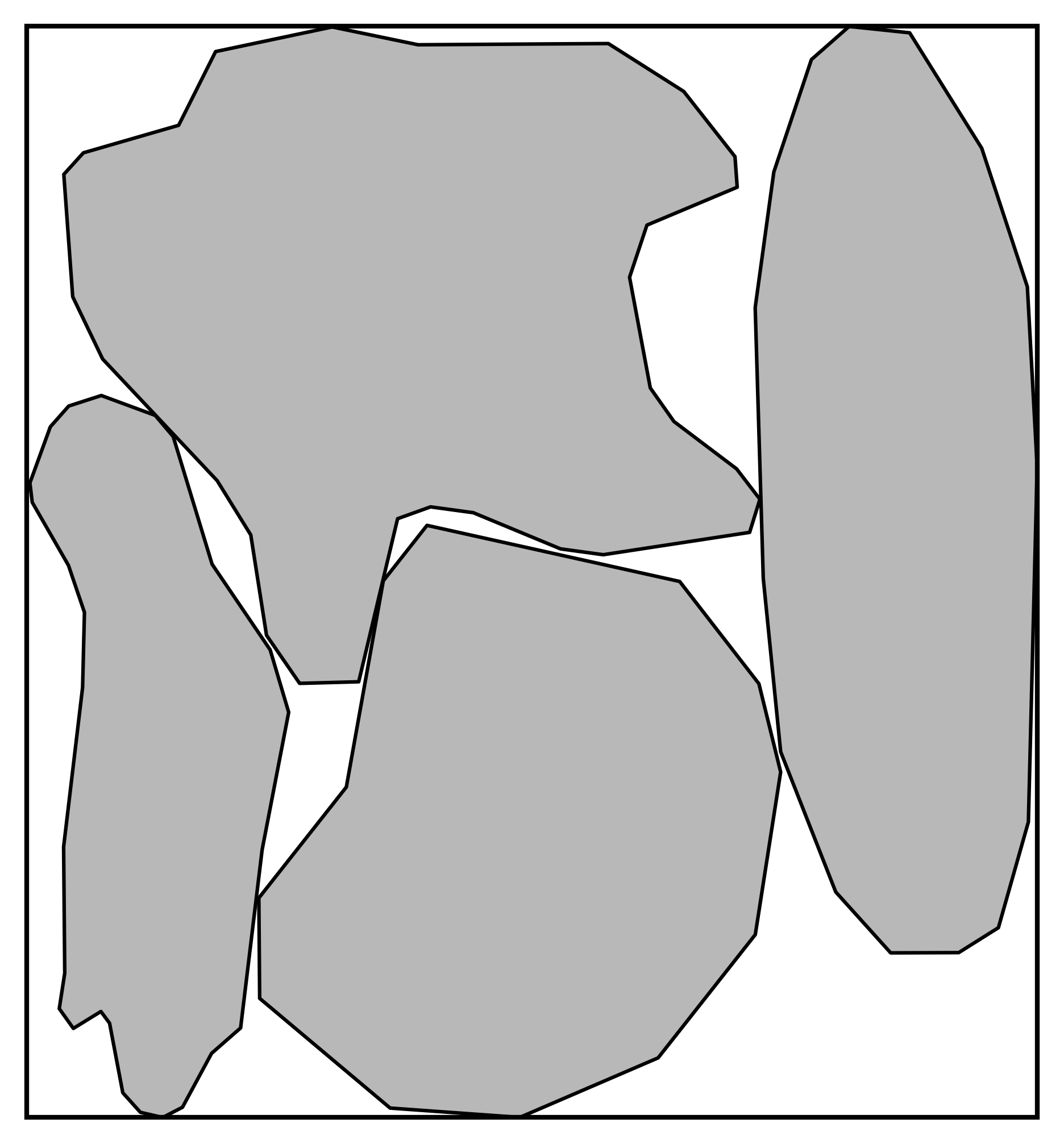}} \\[2pt]
        {\small (a) Compact layout (occupancy $78.0\%$)} &
        {\small (b) Loose layout (occupancy $75.8\%$)}
    \end{tabular}
    \caption{The unconstrained 2D irregular nesting problem: place a set of irregular polygons without overlap so that they occupy the tightest axis-aligned box, with both width and length free.
    A slight change in relative poses can open large empty regions and shift occupancy by several points.}
    \label{fig:problem_def}
\end{figure}

Historically, the Packing community has circumvented the unconstrained problem by focusing on 2D Irregular Strip Packing (2DISPP), utilizing exact solvers that fail to scale due to NP-hardness \citep{MIP}, or heavily tailored heuristics \citep{gomes2006solving}. These classical heuristics remain blind to polygon geometries, relying instead on guided brute-force search to navigate the solution space. While a few machine learning approaches exist \citep{uv_packing, hybrid_rl}, they often struggle with reproducibility and rely on rasterization of the polygons. 

The geometric complexities of unconstrained nesting, specifically ensuring the non-overlap of irregular rotating shapes, make it notoriously difficult to handcraft effective geometric priors. Consequently, current state-of-the-art heuristics \citep{sparrow} achieve their efficiency at the cost of relying on weak geometric priors, navigating the continuous space with minimal structural guidance. We argue that the CORL framework coupled with a tailored neural architecture for polygon encoding can overcome these limitations by automatically learning relevant geometrical priors from the data and guiding the search.

We propose a novel framework that learns to navigate the unconstrained 2D nesting solution space. At the core of our approach is the integration of a Polygons Transformer (PoT) architecture within a CORL training framework. The PoT encoder is specifically designed to capture the geometry of multi-polygons while enabling inter-polygonal attention. By coupling this geometrically aware encoder with an RL agent, we demonstrate that neural combinatorial optimization can successfully tackle combinatorial spatial tasks.

The contributions of this paper are:
\begin{itemize}
    \item \textbf{Formalization for Unconstrained Nesting:} We formulate unconstrained 2D irregular nesting as occupancy-ratio maximization over a free bounding box, a critical industrial setting historically overlooked.

    \item \textbf{Training Dataset \& Benchmark:} To support this new paradigm, we release a dedicated open-source training dataset and an evaluation benchmark, and establish the first Deep Learning baseline, providing the community with a rigorous and reproducible framework for future research.
    
    \item \textbf{Neural Architecture for Polygon Encoding:} We introduce a novel architecture, the Polygons Transformer (PoT), to directly encode 2D continuous vector geometries, allowing exact representation of the shapes and cross-polygon attention.
    
    \item \textbf{CORL Training \& Experimental Validation:} We implement a CORL training pipeline. We empirically show that the learned policy is outperforming Sparrow on 4-polygon instances, and we document a clear scaling gap on 8-polygon instances.
\end{itemize}


\section{Related work}

We review the existing literature across three domains: 2D nesting formulations, polygon representation, and neural combinatorial optimization, highlighting the specific limitations current methods face regarding continuous spatial exploration.

\paragraph{Nesting and Packing Methods} Optimization of spatial arrangements encompasses three progressive formulations: 2D Bin Packing (restricted to rectangular items and boundaries), 2D Irregular Strip Packing (2DISPP, nesting irregular polygons within a fixed-width strip), and unconstrained 2D Irregular Nesting (no predefined boundary constraints). Most existing methods impose bounding containers, restrict rotations to discrete angles, or assume convexity, to mitigate geometric complexity and to satisfy specific industrial constraints. By discarding these limitations, our approach targets a highly specific and rarely addressed formulation: the pure minimal bounding box optimization of irregular, non-convex polygons under free rotation. To the best of our knowledge, only \citet{jones2014fully} treats this exact unconstrained problem. However, this exact mathematical approach struggles to scale when polygon vertices increase due to the problem's inherent NP-hardness. Consequently, we leverage the mature 2DISPP literature, bridging to general nesting by dynamically optimizing the strip width via grid search. The theoretical foundation of this domain traces back to \citet{nfp}, who introduced early bottom-left heuristics and the No-Fit-Polygon (NFP). By geometrically precomputing exact collision boundaries, the NFP avoids complex overlap detection during placement, reducing the continuous action space and remaining a cornerstone of the field. Beyond NFP-based continuous representations, researchers have also tackled geometric complexities using diverse overlap detection methods \citep{sato2019raster, jagua}, while deploying meta-heuristics such as Simulated Annealing \citep{gomes2006solving} and Tabu Search \citep{bennell2001hybridising} to navigate the vast combinatorial placement space. Recently, Sparrow \citep{sparrow} emerged as the state-of-the-art heuristic solver for 2DISPP, providing a robust, open-source benchmark that systematically outperforms classical approaches.

\paragraph{Polygon Encoding} Mapping variable-length vertex sequences into a fixed-dimensional latent space is essential to integrate geometric shapes into Deep CORL frameworks, enabling the state encoder to feed fixed-size representations into the actor-critic decoders. While single-polygon encoding is well-established for supervised tasks through Graph Neural Networks (GNNs) \citep{cgae}, 1D-ResNets and spectral shape descriptors (NUFT) \citep{nuft}, nesting inherently requires multi-polygon representations to capture partially filled containers and spatial relationships. To address this, spectral approaches like NUFT can accommodate multiple shapes, whereas graph-based architectures like PolyGNN \citep{poly_gnn} are explicitly designed for this multi-polygon setting by modeling spatial relationships through cross-polygon visibility edges. More recently, Transformer-based architectures \citep{poly_transformer} have advanced multi-polygon processing by leveraging cross-polygonal vertex attention to efficiently extract geometric relationships between polygons. Despite these representational advances, current Deep Learning approaches applied to nesting predominantly bypass them in favor of rasterization \citep{uv_packing}. This pixel-grid discretization is sub-optimal for our industrial use case, as it incurs severe information loss, sparse representation, and lacks the millimeter-precision strictly required for PCB manufacturing.

\paragraph{Neural Combinatorial Optimization (NCO)} NCO replaces handcrafted heuristics with data-driven policies. Unlike traditional solvers that require extensive re-engineering whenever problem constraints shift, NCO automatically derives general-purpose heuristics from training data, reducing domain-specific design. Early supervised NCO methods successfully imitated expert solvers but generalized poorly to out-of-distribution instances \citep{nco}. Combinatorial Optimization Reinforcement Learning (CORL) \citep{corl} bypasses this limitation by allowing automatic exploration of the solution space. In \citet{RL4CO}, paradigms are typically categorized into Learning to Construct (L2C) \citep{kool2018attention} and Learning to Improve (L2I) \citep{wu2021learning}. In L2C, the agent builds the solution from scratch by iteratively placing each component, whereas in L2I, the agent starts from a complete, initialized configuration and selects which components to perturb. Training typically uses Policy Gradient algorithms (e.g., REINFORCE, PPO), with variance reduction achieved via learned critics or multiple-rollout baselines like POMO \citep{pomo}.

Despite its sample efficiency, L2C suffers from sparse, episode-end rewards that make training a critic notably hard \citep{pomo}. Workarounds like POMO collapse the problem into a one-step MDP for baseline computation, thus abandoning intermediate credit assignment, obscuring individual action contributions, and restricting batch diversity by requiring equivalent state rollouts. L2I overcomes these structural limits by deriving dense rewards from consecutive state improvements. This step-wise signaling enables precise credit assignment and advantage estimation, natively accommodating actor-critic architectures.

Rather than maximizing expected returns, CORL aims to maximize peak performance within an inference budget. Because simulating expensive inference budgets during training is extremely sample-inefficient, the core challenge lies in designing surrogate training rewards that correlate strongly with peak performance. Constructive methods typically try to address this by isolating top trajectories: Poppy \citep{Poppy} backpropagates exclusively on the best performing agent from a small population, while Compass \citep{Compass} optimizes a noise-conditioned meta-policy via its best-performing noise vector, successfully decoupling constrained training batches from expansive inference diversity. Alternatively, L2I naturally supports arbitrarily large inference budgets. The max reward formulation in DACT \citep{dact} seems to better align with the inference objective by exclusively rewarding the discovery of new global optima. Nonetheless, optimizing for the best found configuration across one episode metric might not be correlated with the best found configuration across a budget of thousands to millions of episodes.

Despite CORL's success in routing, its application to 2D irregular nesting remains limited. GFPack++ applies diffusion models as a supervised NCO approach for imitation learning \citep{gfpack}, which inherently restricts its ability to explore beyond its training distribution. Alternatively, RL-based UV Packing \citep{uv_packing} scales to massive instances (e.g., 200 polygons) via a constructive (L2C) approach; however, it relies on rasterization and lacks an open-source implementation for direct comparison. Finally, \citet{hybrid_rl} proposes a hybrid RL algorithm for irregular packing, but it lacks explicit geometric encoding details and strictly constrains rotations to discrete orthogonal angles, preventing fine-grained spatial optimization.

\section{Unconstrained 2D irregular nesting}
\paragraph{Problem Context}
The motivation for defining this specific variant of the nesting problem stems directly from practical challenges encountered in modern manufacturing, specifically within Printed Circuit Board (PCB) production. In this industry, individual PCB outlines must be panelized within a larger rectangular production board. Because these boards are processed on adjustable conveyors, their dimensions are not strictly fixed but instead fall within a broad, continuous operational range.

Additionally, production standards in this sector dictate specific parameters for item count and shape. Manufacturing units typically comprise a small kit of components, rarely exceeding eight polygons. Because end-use applications like wearable electronics and embedded systems demand strict spatial efficiency, the outlines of these PCBs are frequently non-convex and highly irregular, with many vertices. Consequently, we formalize this specific industrial challenge as a 2D Irregular Nesting Problem, focusing on the deep-search optimization of a limited number of complex geometric shapes. This proposed paradigm differs from the standard 2DISPP solely by relaxing the strict fixed-width constraint and restricting the scope to small item sets.

\paragraph{Problem Formulation}
Given $N$ irregular polygons $\{P_i\}_{i=1}^{N}$, a nest assigns each piece a rotation $\theta_i$ and a translation $\mathbf{t}_i$ so that interiors remain disjoint. The panel is the axis-aligned bounding box of the posed pieces: both width $W$ and height $H$ are free. Quality is the occupancy ratio
\[
O = \frac{\sum_{i=1}^{N} \mathrm{area}(P_i)}{WH}.
\]
Rotations are unrestricted, as it can be observed for several industrial use cases. Occupancy gains of a fraction of a percent compound over mass production, so high-quality nests are obtained by allocating a substantial computational budget per instance.

\paragraph{Polygons Dataset}
Training and validating a generalized policy for this domain requires an extensive dataset of irregular polygons with significant geometric diversity. Because conventional polygon databases tend to be biased toward regularity, they fail to capture the complex, non-convex nature required to rigorously test our algorithms. To capture true irregularity, we exploit the natural complexity of terrestrial coastlines and historical borders. We found the ideal data source in the OpenStreetMap (OSM) \citep{OpenStreetMap} land dataset, which catalogs the boundary polygons of all terrestrial land areas globally. Coastlines yield the most irregular polygons we could assemble at scale, and their non-convex, high-frequency contours most closely resemble the manufactured PCB outlines that motivate this work. By utilizing the outlines of continents and islands, we ensure the model is trained on a rich diversity of naturally occurring, highly complex geometries. 

To prepare the raw geographic data for training, we implemented a streamlined preprocessing pipeline. This involved basic geometric simplification, filtering out outliers such as non-closed polygons, discarding overly dense polygons exceeding 200 vertices to maintain computational tractability, and applying shape rescaling to limit the polygon size variance to a factor of 10. Ultimately, this processing extracted a refined dataset of 700,000 unique irregular polygons. 

\paragraph{Benchmark}
To rigorously validate our approach against baselines, we introduce a novel benchmark explicitly designed for the few-polygon 2D Irregular Nesting Problem. The evaluation framework is structured around two distinct complexity tiers: 4-item and 8-item nesting configurations. Each tier comprises 15 unique problem instances which were held out during training to prevent data leakage.

In alignment with open-science principles, this dataset will be self-hosted, maintaining licensing continuity with the original OpenStreetMap data. By releasing both our algorithmic framework and this standardized benchmark, we seek to advance Sparrow's initiative to reinvigorate the nesting open-source research community, focusing on an overlooked, yet significant industrial use case. \footnote{The complete training dataset, benchmark, and source code will be open-sourced upon publication.}

\section{Polygons Transformer (PoT)}

To successfully apply the CORL framework to the 2D irregular nesting problem, the agent must be capable of intelligent spatial exploration guided by strong geometric priors. However, the environment state consists of vector data: a dynamic number of polygons, each defined by a variable-length sequence of vertices. Integrating this state into a deep RL architecture requires an encoder capable of mapping these multi-polygon vector representations into a fixed-dimensional latent space.

\begin{figure}[htbp]
    \centering
    \includegraphics[width=0.8\textwidth]{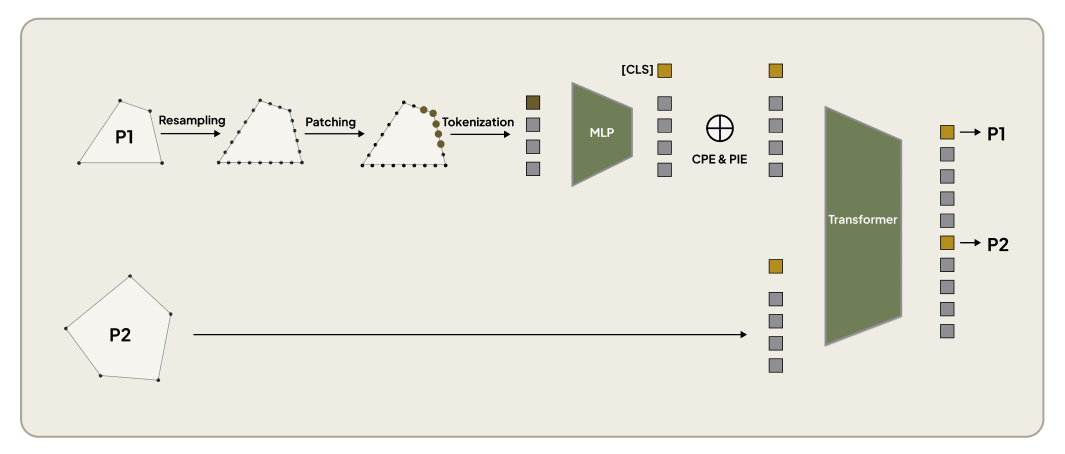}
    \caption{PoT encoding pipeline. Each input polygon's contour is uniformly resampled and grouped into local patches, which are tokenized and projected by an MLP. Tokens are then augmented with cyclic positional embeddings (encoding position within each polygon's closed contour) and polygon ID embeddings (identifying which polygon a token belongs to), before self-attention is applied across the full multi-polygon sequence to yield one geometry-aware embedding per polygon.}
    \label{fig:pot}
\end{figure}

While several learning-based encoders exist, most are restricted to single polygons \citep{cgae, nuft}. Among multi-polygon encoders, spectral approaches like NUFT are non-parametric and lack the flexibility of learned representations, whereas graph-based methods like PolyGNN \citep{poly_gnn} enforce strict rotation invariance. In the context of nesting, rotation invariance is fundamentally incompatible with our objective, as the RL agent must be able to perceive the current absolute orientation of a shape to predict its optimal rotation. Transformer-based tokenization offers a promising alternative. Inspired by the foundational cross-polygonal attention concepts of PSRT \citep{poly_transformer}, we introduce the Polygons Transformer (PoT). To bridge the gap between raw continuous coordinates and actionable state representations, the PoT processes vector geometries through a dedicated three-stage pipeline. As illustrated in Figure~\ref{fig:pot}, the architecture extracts fine-grained local motifs via contour resampling, injects topological context, and distills global spatial relationships through flexible \texttt{[CLS]} aggregation.

\paragraph{Contour Resampling and Patch Tokenization} As the first step of our encoding pipeline, rather than directly tokenizing raw, unevenly spaced vertices, we uniformly resample the contour of each polygon at a higher frequency. This operation serves a dual purpose: it ensures a consistent geometric resolution across all shapes regardless of their original sampling density, and it provides the fine-grained token-level granularity necessary for the RL actor to make highly precise placement decisions later in the process. Furthermore, to process these vertices efficiently, the PoT groups sequential vertices into local geometric patches. Inspired by the ViT architecture \citep{vit}, this tokenization per patch effectively compresses the sequence length, reducing the quadratic computational bottleneck of the Transformer while explicitly capturing local geometric motifs such as edges and corners.

\paragraph{Structural and Geometric Embeddings}
To enable cross-polygonal interactions, the individual patch sequences of all polygons within the environment are flattened and concatenated into a single, unified 1D input sequence. However, this flattening strips away the spatial boundaries between shapes. To provide the Transformer layers with the necessary topological context, the embedded patches are augmented with two specific positional signals before processing:
\begin{itemize}
\item \textbf{Cyclic Positional Embedding (CPE):} Standard linear positional encodings break the closed-loop topology of a polygon. To preserve the structural boundary, we inject cyclic positional embeddings at the patch level. For a patch at position $p$ within a polygon contour consisting of $L$ total patches, we map its position to an angle $\theta_p = 2\pi \frac{p}{L}$. We draw inspiration from Fourier feature networks \citep{tancik2020fourier}: $v_{p} = \bigoplus_{k=0}^{K-1} [\sin(2^k \theta_p), \cos(2^k \theta_p)]$.

This concatenated feature vector $v_{p}$ is then linearly projected to match the Transformer's embedding dimension. This mechanism ensures the network inherently understands that the first and last patches of a shape are contiguous.
\item \textbf{Polygon ID Embedding (PIE):} Because multiple polygons are merged into the same long sequence, the network must be able to distinguish between separate shapes. We add a learned polygon identifier embedding to each token, explicitly encoding which patch belongs to which specific polygon.
\end{itemize}

\paragraph{Cross-Polygon Attention and Flexible Aggregation} The augmented token sequence is passed through a series of standard Transformer self-attention blocks \citep{attention}. This mechanism is highly synergistic with the nesting problem: attention heads learn to model internal polygon geometry using the cyclic positional context, while simultaneously leveraging the multi-polygonal sequence to perform cross-polygon attention and detect geometrical relationships between distinct shapes. Finally, to bridge the variable-length sequence output to the fixed-size input required by the CORL actor-critic networks, we add a dedicated \texttt{[CLS]} token for each polygon within the sequence. 
Functioning as a dedicated latent descriptor for each shape, the \texttt{[CLS]} tokens leverage the global self-attention map to aggregate both intrinsic geometric motifs and inter-polygon spatial relationships. The resulting fixed-dimensional embeddings provide a representation of each polygon, supplying the downstream RL agent with the exact structural context required to navigate the action space.

\section{CORL for 2D irregular nesting}
\label{sec:sec5}

\paragraph{Learning Framework} The objective of nesting is to find a configuration that maximizes the spatial occupancy of a given set of polygons. Motivated by the structural advantages outlined in the related work, we frame our problem as an improvement optimization and train an actor-critic policy with Proximal Policy Optimization (PPO) \citep{schulman2017ppo}. At inference, L2I-trained agents can be used to refine best-found configurations, making L2I efficient for finding the optimal solution when the budget is high. We consider two reward formulations: the dense reward and the max reward. Our default is the dense reward, given at each step as the difference in occupancy ratio between the current step and the previous step:

$$R(a_t, s_t) = O(s_{t+1}) - O(s_t) \quad \text{with $O$ the occupancy ratio of the panel.}$$

We also evaluate the max reward formulation \citep{dact}, which eliminates penalties for degrading the current configuration and is intended to push agents toward the absolute peak solution of a given instance:
$$R(a_t, s_t) = \max(0, O(s_{t+1}) - O_{\text{peak}<t}).$$

\paragraph{Markov Decision Process}
We cast nesting improvement as a deterministic Markov Decision Process. The state $s_t$ comprises the current panel, the $N$ posed polygons, together with the normalized step index and $\rho^{\mathrm{best}}_t$, the highest occupancy observed in the rollout. These scalars preserve the Markov property and tell the critic which peak the policy must exceed to collect a positive reward. An action relocates a single piece,
$a_t = \big(i,\,\Delta\theta,\,\mathbf{t}\big)$,
selecting polygon $i$, applying a relative rotation $\Delta\theta \in [-\pi,\pi]$, and translating its centroid by $\mathbf{t} \in \mathbb{R}^2$. Transitions are a deterministic function of $(s_t, a_t)$.
The learned policy acts on a discretization of this interface. Translation is restricted to the NFP boundary of the moving piece against the stationary others, itself discretized into patch tokens $j$ and subregions $s$. Rotation is quantized into $K=720$ uniform bins of $0.5^\circ$. The resulting discrete action is $a_t = \big(i,\,k,\,j,\,s\big)$.

\paragraph{Why Discrete Rotation.}
A continuous Gaussian rotation head could in principle emit any angle in $[-\pi,\pi]$. We instead use a categorical over a fine grid. \emph{Multimodality} is valuable during exploration: several distant angles can be locally excellent, and the policy should keep those candidates in memory rather than collapsing onto a single mode. A univariate Gaussian cannot do this: it is unimodal and symmetric, a shape with no geometric justification. A mixture of Gaussians can, but only after fixing the number of components in advance, and mixture policies are notoriously unstable under policy gradients. A categorical over $720$ bins has no assumed density shape and can place mass on an arbitrary number of modes; at $0.5^\circ$ the grid is dense enough to approximate the continuum. We lock \texttt{Sparrow} and \texttt{SVGNest} to the same resolution so the comparison is not confounded by angular budget.

\paragraph{NFP-guided Placement.}
A core insight of our approach is reducing the two-dimensional translation problem to a one-dimensional search. For a moving polygon $P_i^{\theta}$ ($P_i$ rotated by angle $\theta$), the forbidden region created by stationary obstacles $P_j$ is the union: $\mathcal{F}(i,\theta) = \bigcup_{j \neq i} \mathrm{NFP}\!\left(P_j, P_i^{\theta}\right).$

For any high-quality configuration, any polygon lies on the boundary $\partial \mathcal{F}(i,\theta)$ of the other polygons in the panel. If the reference point lies in the interior of $\mathcal{F}(i,\theta)$, the polygons overlap; if it lies strictly outside, the layout quality can be improved or matched by translating the piece until it first contacts another, which places the reference point on $\partial\mathcal{F}(i,\theta)$. Therefore, searching for optimal placement can be reduced to searching on the boundary of the union of NFP polygons. Notably, this boundary may contain holes, reflecting disjoint regions of valid placements. As detailed in the PoT Encoder section, we uniformly sample this continuous boundary. This discretization enables the categorical policy to select discrete translation coordinates that guarantee collision-free placements by construction.

\paragraph{Two-Stage Actions.}
Factoring selection, rotation, and placement into two stages is heavily computationally motivated. Stage~1 decides \emph{which} polygon to move and \emph{how} to rotate it based entirely on the current panel configuration. The agent predicts a rotation distribution for all polygons; rotation is then sampled from the distribution of the selected polygon. Stage~2 then decides \emph{where} to place it conditionally on polygon selection and rotation. Because feasible boundaries depend strictly on the selected polygon's identity and absolute orientation, NFPs are computed on-demand only after Stage~1 completes. The joint action log-probability decomposes additively:
\begin{equation}
    \log \pi(a_t) = \log \pi_{\mathrm{item}}(i) + \log \pi_{\mathrm{rot}}(k \mid i) + \log \pi_{\mathrm{place}}(j,s \mid i,k).
\end{equation}
This decoupling shields Stage~1 from expensive geometric computations (computing all possible NFPs) and prevents conditioning the selection policy on undefined geometry, all while remaining end-to-end differentiable.

\paragraph{Observation Space.}
The dynamically computed NFP patch tokens are excluded from the primary panel observation and fed exclusively to Stage~2. This isolation keeps the main panel encoder stable across steps, allowing the Stage~2 placement network to specialize entirely on instance-specific feasible boundaries.

\begin{figure}[htb]
    \centering
    \includegraphics[width=1\textwidth]{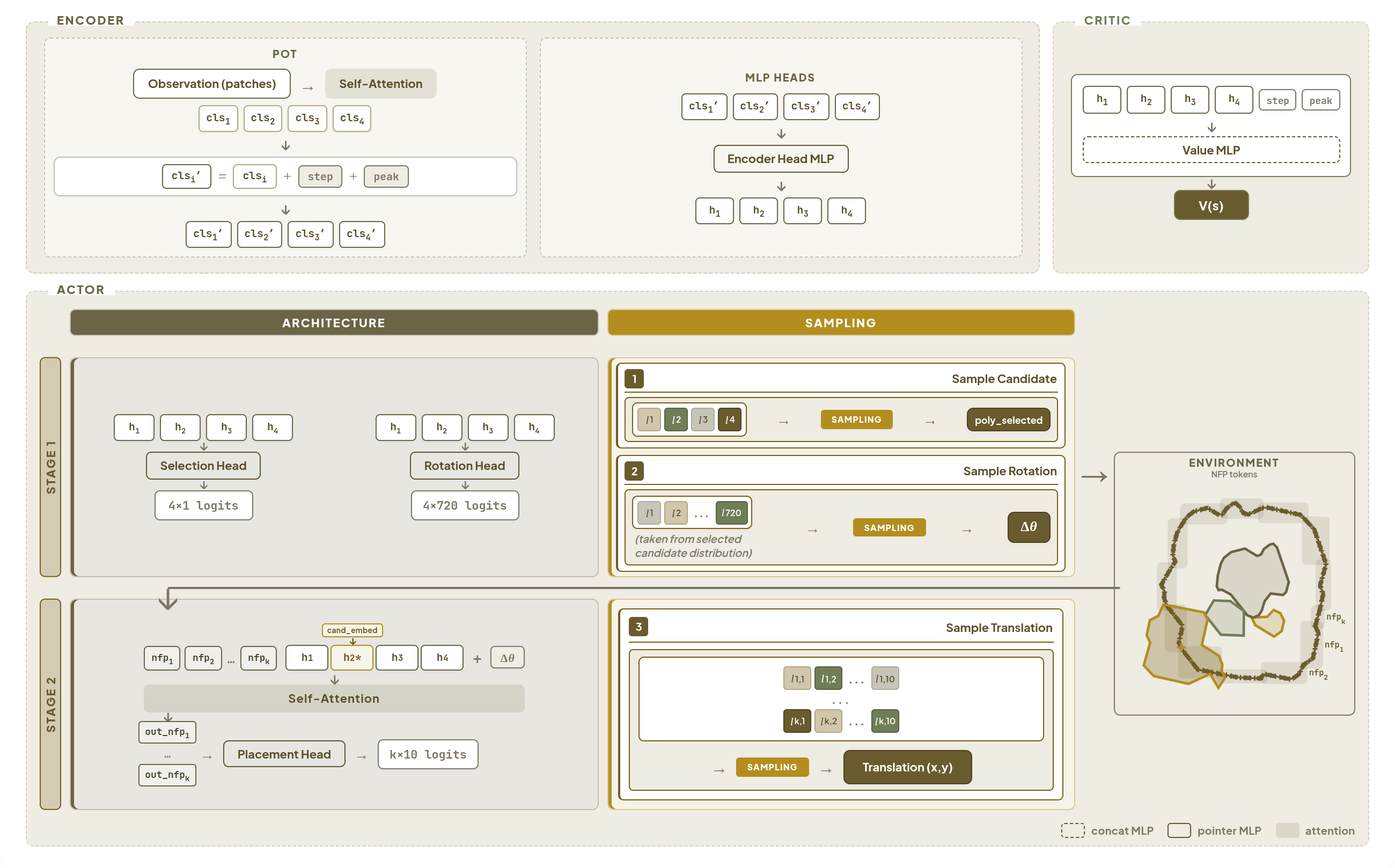}
    \caption{Actor-critic architecture of the nesting agent.
    PoT (Figure~\ref{fig:pot}) produces cross-attended polygon embeddings, which we enrich with the step index and previous peak occupancy (the latter keeps max-reward training Markovian).
    A critic head maps these tokens to a value $V(s)$.
    The actor first samples the candidate piece and the rotation to apply; the environment then removes and rotates that piece and returns the NFP as a 1D feasible contour.
    Stage~2 attends over this contour, the remaining placed polygons, and the candidate to predict a distribution over NFP-patch vertices at which to place the candidate.}

    \label{fig:architecture}
\end{figure}

\paragraph{Neural Architecture.}
Figure~\ref{fig:architecture} illustrates the complete agent architecture. Resampled panel contours are patchified and processed by a PoT backbone alongside a per-polygon index embedding. This encoder outputs one CLS token per polygon, which is augmented with normalized step and peak occupancy scalars and passed through an MLP to generate the geometric representation vectors $\mathbf{h}_1,\ldots,\mathbf{h}_N$ for the $N$ polygons. 

The critic network flattens these representation vectors alongside the occupancy scalars, mapping them to a baseline value estimate via an MLP. Simultaneously, Stage~1 processes the representation vectors in parallel using shared heads to produce selection and discrete rotation logits per polygon. The policy samples these logits to determine the target polygon $i$ and its relative rotation $\Delta\theta$.

Given the selected action tuple $(i, \Delta\theta)$, the environment dynamically computes the NFP boundary union $\partial \mathcal{F}(i,\Delta\theta)$. This one-dimensional boundary is resampled, patchified, and augmented with explicit embeddings to distinguish exterior boundaries from holes. 

Finally, Stage~2 combines the NFP patch tokens, a rotation angle embedding ($\Delta\theta$), and the candidate polygon representation vector ($\mathbf{h}_1,\ldots,\mathbf{h}_N$) into a single sequence. This concatenated sequence is processed by self-attention blocks before a pointer-style placement head outputs logits over each NFP token. These logits map to subregions of the NFP patch so that the placement head produces a distribution over NFP contour vertices.

\paragraph{Inference Strategy} 
A key advantage of the L2I framework is its ability to initialize with high-quality placements, allowing the improvement agent to refine them into superior configurations. This mechanism is central to our inference strategy: we maintain a buffer of the top-performing placements and periodically sample from it to establish an episode's initial state. As demonstrated in Table~\ref{tab:buffer_strategy} in the Appendix, our most effective solutions were predominantly achieved by iteratively improving upon these already-strong configurations.

\section{Experiments}
\label{sec:experiments}

In this section, we evaluate the performance of our proposed Polygons Transformer (PoT) framework against a state-of-the-art baseline adapted for unconstrained 2D irregular nesting. We first outline our experimental setup, then present a comparative performance analysis, and conclude with ablation studies.

\subsection{Experimental Setup}
\label{subsec:setup}

\paragraph{Benchmark and Evaluation.} We evaluate our model on the novel small-scale 2D unconstrained irregular nesting benchmark introduced in Section 3, utilizing both the 4-item (P4) and 8-item (P8) complexity tiers. For each method and tier, we run three independent random seeds, compute the mean best occupancy ratio over all 15 instances for each seed, and report the mean and population standard deviation of those three seed-level benchmark scores.

\paragraph{Model Selection.} We evaluate the checkpoint that attains the highest average best occupancy per episode during training. We do so because, for a given run, this one-episode metric is significantly correlated with the average best occupancy over $10^6$ episodes, the inference metric, as shown in Figure~\ref{fig:train_eval_occ} of the Appendix. This criterion cannot be used for the max-reward run: once the best occupancy per episode is itself the training reward, the correlation no longer holds. For that run we simply evaluated several checkpoints under the inference budget and retained the best.

\paragraph{Baseline Configuration.} As there is no established baseline tailored specifically for unconstrained 2D irregular nesting, we adapt \texttt{Sparrow}, the current SOTA heuristic for the 2DISPP \citep{sparrow} and \texttt{SVGNest}, the most starred 2DISPP baseline on GitHub \citep{svgnest}. Because both solvers already search a discrete orientation set, we lock their angular resolution to $0.5^\circ$ ($720$ uniform bins), matching the categorical rotation grid of our policy. To optimize both layout length and width simultaneously, we wrap the core engine of \texttt{Sparrow} and \texttt{SVGNest} in a grid-search layer over the container dimensions. We provide the comprehensive algorithmic details in Appendix~\ref{app:baseline_details}.

\paragraph{Implementation and Training Details.} Our framework is implemented in PyTorch and trained with PPO on 8 NVIDIA RTX 6000 Blackwell GPUs for approximately $240$ hours per tier, for 10 billion steps. To optimize learning stability and throughput, our pipeline incorporates two key training enhancements. First, we employ a curriculum learning strategy adapted from \citet{dact}. Training episodes are set to 80 steps, but we initialize each episode by rolling out the current policy for 20 preliminary steps, computing network gradients exclusively on the remaining 60 steps. This exposes the model to harder configurations, avoiding obvious steps. Second, we implement advantage filtering \citep{gigaflow} to discard low-impact samples. Transitions yielding an absolute Generalized Advantage Estimate (GAE) below a dynamic threshold $\eta$ are skipped during backward passes. Empirically, this filters out between $20\%$ and $40\%$ of all transitions, improving the training sample efficiency. Detailed hyperparameter configurations are provided in Appendix~\ref{app:hyperparameters}.

Training is a one-time cost, paid once and amortized over every subsequent instance. At inference the two methods operate in a comparable cost regime: \texttt{Sparrow} runs for 1~hour on 364~CPUs, while PoT runs for 10~hours on 8~GPUs; at typical cloud prices these budgets fall in the same ballpark. Hardware specifications are given in Appendix~\ref{app:experimental_details}.

\subsection{Main Results and Performance Comparison}
\label{subsec:main_results}

We aggregate the performance of our model against the adapted \texttt{Sparrow} and \texttt{SVGNest} baselines across all 30 evaluation instances. Table~\ref{tab:main_results} provides a compact summary of the average occupancy for both complexity tiers. The comprehensive, instance-by-instance metrics are provided in Appendix~\ref{app:experimental_details}.

\begin{table}[H]
\centering
\caption{Mean best area utilization (\%) on P4 and P8 for the three methodologies. Values are mean $\pm$ population standard deviation over 3 independent seeds.}
\label{tab:main_results}
\setlength{\tabcolsep}{8pt}
\renewcommand{\arraystretch}{1.08}
\begin{tabular}{@{}l ccc@{}}
\toprule
\textbf{Test set} & \textbf{PoT (ours)} & \textbf{Sparrow} & \textbf{SVGNest} \\
\midrule
P4 (4 polygons)  & $\mathbf{81.18 \pm 0.02}$ & $81.07 \pm 0.02$ & $72.28 \pm 0.11$ \\
P8 (8 polygons)  & $81.79 \pm 0.15$          & $\mathbf{85.19 \pm 0.19}$ & $68.59 \pm 0.30$ \\
\bottomrule
\end{tabular}
\end{table}

\paragraph{Analysis of Performance}
As summarized in Table~\ref{tab:main_results}, our learned geometric policy slightly outperforms the \texttt{Sparrow} baseline on the 4-polygon tier. PoT shows a $+0.11$ occupancy-point mean gap, with a $10/5$ win/loss record across the 15 instances (Table~\ref{tab:p4_all_seeds}). Seed-to-seed variation is negligible on this tier ($\pm 0.02$ for both methods), so the ranking is stable. Managing to slightly outperform such a heavily optimized heuristic is a significant milestone for a learned policy, indicating that the PoT encoder extracts functional geometric priors from raw vector data without human-engineered placement rules.

On the more challenging 8-polygon tier, Sparrow is the strongest method, securing a $3.40$ occupancy-point gap over PoT and winning on $13$ of the $15$ instances. Because architecture and hyperparameters were tuned primarily on P4, this deficit maps a clear scaling gap of the current policy. Run-to-run variance also grows with instance size, but it remains small compared with the method gap.

Relative to both, SVGNest lags substantially, trailing PoT by $8.90$ and $13.20$ occupancy points on P4 and P8 respectively. The gap is unambiguous: a classical NFP + genetic-algorithm engine, even with the same two-stage container search, remains well below both the SOTA strip-packing heuristic and the learned policy on this unconstrained bounding-box task.

\subsection{Ablation Studies}
\label{subsec:ablation}

To isolate the contribution of each of our core design choices, we conduct a series of controlled ablation studies on the P4 tier. Starting from our full configuration (2-Step Discrete NFP placement, Dense Reward formulation, and 20-step curriculum), we independently vary a single component at a time while holding all other hyperparameters fixed. Each variant is evaluated over three inference seeds; Table~\ref{tab:ablation} reports the mean occupancy and the population standard deviation of those three seed-level scores, matching Section~\ref{subsec:setup}.

\begin{table}[htbp]
\centering
\caption{Ablation study on the P4 tier over 3 inference seeds. Occupancy is the mean best area utilization (\%); values are mean $\pm$ population standard deviation over the three seed-level scores. $\Delta$ is relative to the default mean.}
\label{tab:ablation}
\begin{tabular}{@{}llcc@{}}
\toprule
\textbf{Ablation Axis} & \textbf{Configuration} & \textbf{Mean $\pm$ std} & \textbf{$\Delta$} \\
\midrule
-- & \hspace*{-0.9em}Default (2-Step NFP + Dense Reward + Curriculum) & $81.18 \pm 0.02$ & -- \\
\midrule
Reward Formulation & Max Reward & $80.63 \pm 0.06$ & $-0.55$ \\
Placement Architecture & 1-Step Gaussian & $81.09 \pm 0.00$ & $-0.09$ \\
Curriculum & No Curriculum (0 steps) & $80.93 \pm 0.05$ & $-0.25$ \\
\bottomrule
\end{tabular}
\end{table}
\paragraph{Dense vs.\ Max Reward.} Replacing Dense Reward with the Max Reward formulation of \citet{dact} yields $80.63\%$, a $0.55$ occupancy-point drop and the largest gap in Table~\ref{tab:ablation}. We return to this result in the discussion of surrogate rewards below.

\paragraph{2-Step Discrete NFP vs.\ 1-Step Gaussian.} We next evaluate the relevance of our two-stage decomposition (Section~\ref{sec:sec5}), which first selects a polygon and rotation, then samples a placement from the NFP-constrained boundary. As a naive single-stage alternative, we implement a 1-Step Gaussian architecture that jointly regresses a rotation angle and a continuous $(x, y)$ centroid translation target from a Gaussian policy head, without explicit NFP conditioning. To preserve feasibility, the predicted centroid is subsequently projected (``scraped'') onto the nearest point of the NFP boundary induced by the other stationary polygons at the predicted rotation. This 1-Step Gaussian baseline reaches $81.09\%$, $0.09$ occupancy points below the default, and its three-seed range does not overlap the default range.

\paragraph{Curriculum vs.\ No Curriculum.} Finally, we ablate the curriculum learning strategy adapted from \citet{dact} (Section~\ref{subsec:setup}), which rolls out the policy for 20 preliminary, gradient-free steps before beginning backpropagation on the remaining episode steps. Removing the curriculum (0 preliminary steps) yields $80.93\%$, a $0.25$ occupancy-point drop relative to the default configuration, confirming that exposing the agent to harder, further-from-initial configurations via the curriculum provides a measurable benefit, consistent with its intended role in the L2I framework.

\paragraph{Discussion.} For every ablated axis, the three-seed occupancy range of the variant does not intersect the default range, so the ranking is stable despite small absolute gaps. The 2-Step Discrete NFP architecture, the Dense Reward formulation, and the curriculum strategy all outperform their respective alternatives, supporting their inclusion in our default configuration. We do not ablate the panel encoder itself: Stage~2 is a pointer network over NFP contour tokens, so the backbone must ingest a variable-length multi-polygon state and attend to every candidate placement vertex. Raster or single-polygon encoders cannot implement this placement head.

\paragraph{On Surrogate Rewards and Reward Hacking.}
Max Reward was introduced in Section~\ref{sec:sec5} because it appears closer to the inference objective than Dense Reward: it credits new occupancy peaks rather than step-to-step changes. Replacing Dense Reward with this formulation nevertheless costs $0.55$ occupancy points (Table~\ref{tab:ablation}). A natural explanation would be that one-episode peak occupancy, the quantity Max Reward optimizes, simply does not predict performance after $10^6$ inference episodes. That is not the case. Under Dense Reward, the training checkpoints with the highest one-episode peak are also those that perform best when later evaluated for $10^6$ episodes (Figure~\ref{fig:train_eval_occ}). When Max Reward is the training objective, this agreement disappears: a checkpoint with a higher one-episode peak can yield a lower occupancy once evaluated over $10^6$ episodes. A quantity that correlates with the intended objective is therefore not automatically a sound training reward, since optimizing it can undo the correlation that motivated its use. An extended analysis is given in Appendix~\ref{app:train_eval_corr}.

\section{Conclusion and Future Work}
\label{sec:conclusion}

We formalized unconstrained 2D irregular nesting and introduced PoT, an attention-based multi-polygon encoder trained within a L2I framework. Without hand-designed placement rules, the resulting agent matches, and slightly exceeds, the state-of-the-art \texttt{Sparrow} heuristic on the 4-polygon tier, and clearly outperforms the genetic \texttt{SVGNest} baseline across all tested scales. Our ablations support our design choice: two-stage NFP-guided placement outperforms direct continuous regression, and curriculum learning outperforms its absence. More surprisingly, our simple Dense Reward outperforms the Max Reward of \citet{dact}.

This last finding points to our main limitation. As shown in Appendix~\ref{app:train_eval_corr}, a cheap one-episode proxy tracks inference performance within a given training run, but this relationship weakens across runs and collapses once the same quantity is used as a surrogate reward. Without a training surrogate that remains faithful under optimization, architectural and reward changes must be validated at full inference cost, which slows iteration on our architecture. A second limitation appears at scale: on the 8-polygon tier, \texttt{Sparrow} regains a clear advantage. This partly reflects design choices tuned for smaller instances. It may also reflect the scope of the action: each transition relocates a single piece while the others stay frozen, so a compact nest of many polygons, which depends on coordinated relative poses, can be reached only through a long sequence of one-piece updates. We did not ablate the neighborhood of the operator, so this reading of the P8 gap remains a hypothesis.

A natural next step is to treat surrogate-reward design as a CORL problem in its own right. The failure of a well-motivated training signal to remain faithful under optimization is unlikely to be nesting-specific; repeating the correlation analysis of Appendix~\ref{app:train_eval_corr} on canonical tasks such as TSP and CVRP, where solvers, budgets, and evaluation protocols are already shared, would isolate the limitation more cleanly and invite further work on methods for testing surrogate fidelity. Scaling to dozens of polygons is a separate question. A joint operator over several pieces at once is one candidate, for example a diffusion-style update, but it remains exploratory until a training surrogate tracks the inference objective.

\newpage

\bibliographystyle{unsrtnat} 
\bibliography{preprint}

\newpage
\clearpage
\appendix
\addcontentsline{toc}{section}{Appendix}

\begin{center}
    {\LARGE \textbf{Appendix}}
\end{center}
\vspace{0.5cm}

This supplementary material provides additional technical details, experimental configurations, and granular baseline evaluation metrics to support the reproducibility of the main manuscript.

\section{Hyperparameters and Architecture Details}
\label{app:hyperparameters}

This section details the explicit parameters used to initialize and train the Polygons Transformer (PoT) encoder and the Combinatorial Optimization Reinforcement Learning (CORL) policy network via PPO. 

Table~\ref{tab:training-hps} provides the complete list of training scales, optimization settings, discrete environment discretization setups, and localized structural configurations.

\begin{table}[H]
\centering
\footnotesize
\caption{Training and model hyperparameters.}
\label{tab:training-hps}
\setlength{\tabcolsep}{4pt}
\renewcommand{\arraystretch}{0.88}
\begin{tabular}{@{}llr@{}}
\toprule
\textbf{Component} & \textbf{Hyperparameter} & \textbf{Value} \\
\midrule
\multicolumn{3}{@{}l}{\textbf{Training}} \\
\cmidrule(lr){2-3}
Scale
  & Total timesteps       & $10^{10}$ \\
  & Parallel envs         & 128 \\
  & Steps per episode     & 80 \\
  & Episodes per env     & 3 \\
  
Optimizer
  & Learning rate         & $10^{-4}$ \\
  & LR Linear annealing          & \checkmark \\
  & Adam $\varepsilon$    & $10^{-5}$ \\
Settings
  & BF16                  & \checkmark \\
PPO
  & GAE $\lambda$         & 0.95 \\
  & Minibatch size        & 5000 \\
  & Update epochs         & 5 \\
  & Advantage norm.       & \checkmark \\
  & Clip coefficient      & 0.05 \\
  & Value clipping        & \checkmark \\
  & Item entropy coef.    & 0.05 \\
  & Rotation entropy coef.& 0.001 \\
  & Placement entropy coef.& 0.01 \\
  & Max grad norm         & 0.5 \\
  & Target KL             & 0.02 \\
  & Discount $\gamma$     & 0.98 \\
  & Value coef.           & 0.2 \\
  & Advantage Filtering factor $\eta$            & 0.02 \\
\midrule
\multicolumn{3}{@{}l}{\textbf{Model}} \\
\cmidrule(lr){2-3}
Encoder
  & Patch size            & 10 \\
Transformer
  & Embed dim             & 128 \\
  & Attention heads       & 4 \\
  & Layers                & 6 \\
  & Fourier bands         & 32 \\
  & LayerNorm $\varepsilon$ & $10^{-5}$ \\
  & Activation            & GELU \\
MLP heads
  & Layers                & 5 \\
  & Hidden dim            & 128 \\
Policy
  & Pose bins             & 720 \\
  & Subregion bins        & 10 \\
  & Stage-2 blocks        & 4 \\
Value head
  & Hidden dims           & 256--128--128--128 \\
\bottomrule
\end{tabular}
\end{table}

\section{Training Metrics vs.\ Inference Performance}
\label{app:train_eval_corr}

The true metric of a nesting policy is its ability to find a high-occupancy configuration under a substantial inference budget. In this work that budget is $10^6$ episodes of $80$ steps each. Running this evaluation at every training checkpoint is prohibitively expensive, so we look for a proxy that remains tractable during training and still indicates whether a model is improving.

The peak occupancy observed over a single training episode is a natural candidate. It is free to compute: the environment already records occupancy at every step, and taking the episode-wise maximum adds no extra rollouts. The open question is whether this one-episode peak is a faithful proxy for the $10^6$-episode inference metric.

\begin{figure}[H]
\centering
\includegraphics[width=\textwidth]{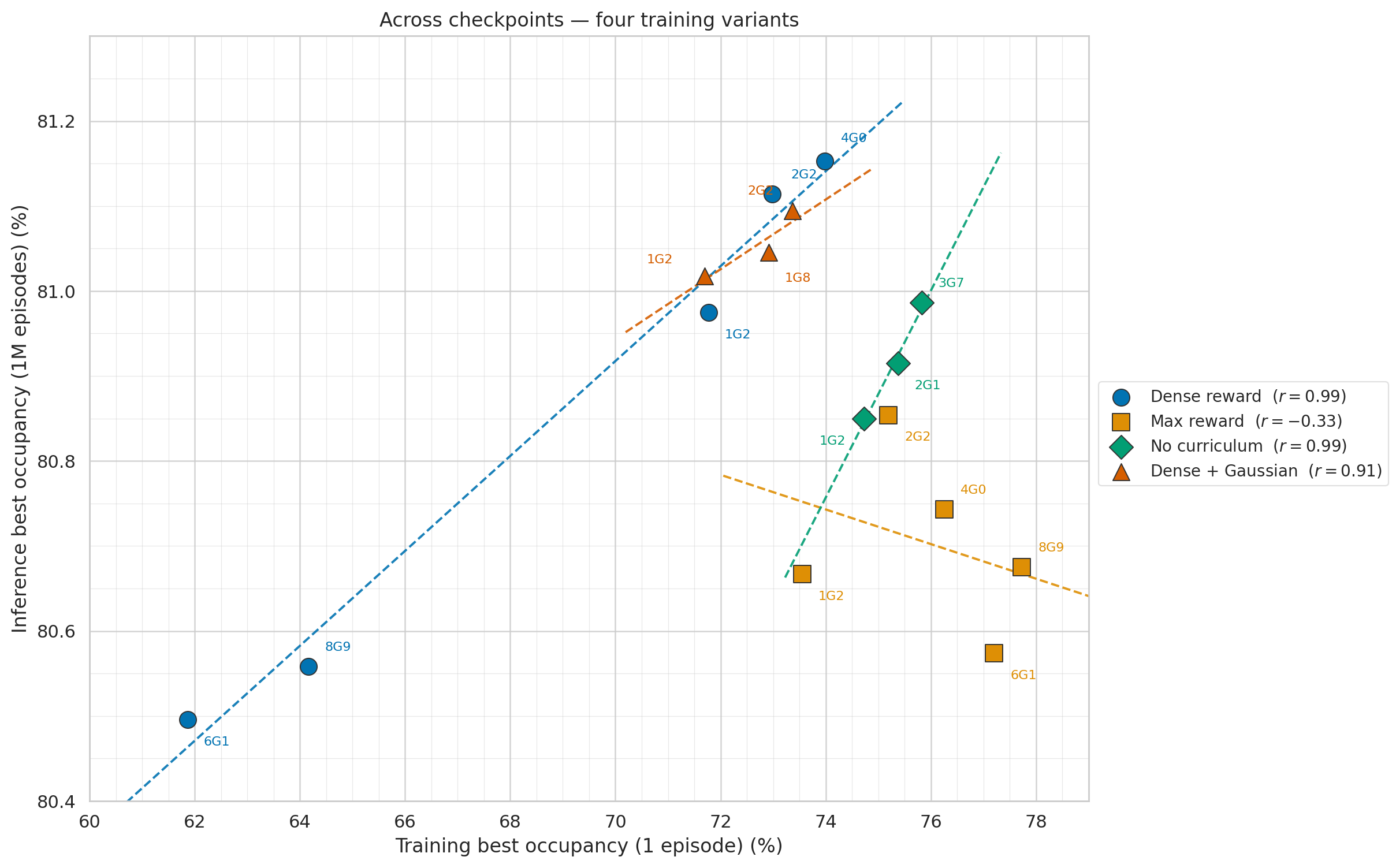}
\caption{Within-run relationship between training 1-episode mean best occupancy and $10^6$-episode inference occupancy on P4.
Dense Reward runs recover the inference ranking of their saved checkpoints ($r = 0.99$ dense; $r = 0.99$ no curriculum; $r = 0.91$ 1-step Gaussian).
Max-reward training, which directly optimizes the one-episode peak, yields $r = -0.33$.}
\label{fig:train_eval_occ}
\end{figure}

Figure~\ref{fig:train_eval_occ} shows that it is a reliable \emph{within-run} checkpoint selector, provided the policy is not trained to optimize that same one-episode quantity. Under the dense reward, ranking the saved checkpoints of a given run by training peak occupancy recovers the $10^6$-episode ranking almost exactly ($r = 0.99$). The same holds for the other dense-reward recipes (no curriculum $r = 0.99$; 1-step Gaussian $r = 0.91$). Under the max-reward formulation, whose explicit goal is to improve the best configuration found within a single episode, the within-run correlation turns negative ($r = -0.33$). Optimizing the proxy itself therefore breaks the relationship that made it useful.

\begin{figure}[H]
\centering
\includegraphics[width=0.72\textwidth]{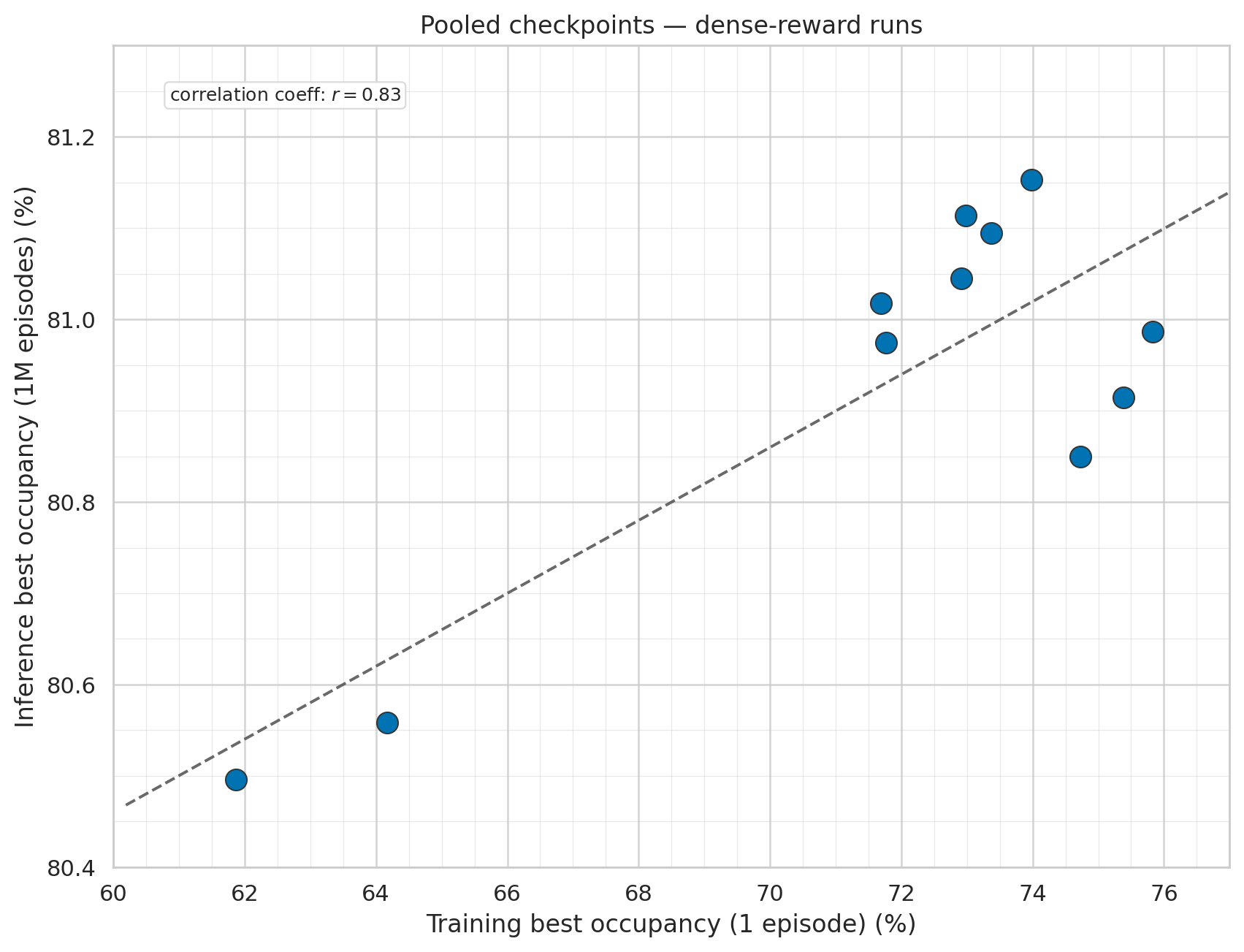}
\caption{The same axes with all dense-reward checkpoints pooled across runs (default, no curriculum, 1-step Gaussian).
The mixed cloud is only moderately correlated ($r = 0.83$), so the one-episode peak is not a score for comparing runs.}
\label{fig:train_eval_occ_mixed}
\end{figure}

The proxy does not extend to comparing distinct runs. Figure~\ref{fig:train_eval_occ_mixed} treats every evaluated checkpoint of the three dense-reward trainings as a point in a single cloud. The mixed association is only moderate ($r = 0.83$): a higher 1-episode occupancy no longer identifies the better model. The one-episode maximum is therefore not a global score of inference performance; it only indicates which of the checkpoints saved in a given run is the one to evaluate.

\section{Baseline Adaptations: Unconstrained Sparrow and SVGNest Setup}
\label{app:baseline_details}

To evaluate our model against competitive classical solvers, we adapt two open-source 2DISPP engines to the unconstrained bounding-box setting: \texttt{Sparrow} \citep{sparrow}, the current state-of-the-art heuristic, and \texttt{SVGNest} \citep{svgnest}, the most starred genetic NFP baseline on GitHub (we run the scriptable Deepnest implementation of the same engine). To the best of our knowledge, the only existing literature addressing the exact unconstrained bounding-box formulation we target relies on exact mathematical programming \citep{jones2014fully}, which becomes computationally intractable for non-convex geometries with complex, high-frequency vertex counts. Furthermore, while alternative reinforcement learning methods like UV Packing \citep{uv_packing} explore similar domains, the lack of an open-source implementation or public code availability prevents direct replication, highlighting the necessity of establishing this baseline framework.

Both engines were designed for strip packing: a container with one free dimension and one fixed dimension. Bridging them to unconstrained 2D irregular nesting therefore requires treating the container dimensions as dynamic optimization variables. We apply the same high-level adaptation to both: lock the discrete orientation grid to match our model, then wrap the core placement engine in a geometry-driven, two-stage grid search over candidate container sizes. The Sparrow instantiation of this wrapper is specified below; \texttt{SVGNest} uses the same protocol with the differences noted at the end of this section.

\subsection{Action Space and Rotation Synchronization}
To ensure a fair evaluation, we synchronize the structural constraints of the baseline with the categorical rotation grid of our policy. Both \texttt{Sparrow} and \texttt{SVGNest} already enumerate a discrete orientation set; we lock that set to $720$ uniform bins of $0.5^\circ$, matching the action space described in Section~\ref{sec:sec5}.

\subsection{Two-Stage Grid Search for Height Optimization}
Without prior layout information, the search space for a viable container height is infinitely large. We constrain and optimize this space using a sequential coarse-to-fine strategy:

\paragraph{Geometric Bound Initialization} 
We initialize the search boundaries directly from the geometry of the target instance. For a given set of $N$ polygons, we calculate the maximum diameter $d_i$ (the longest Euclidean distance between any two vertices on the contour) for each shape. We sort these diameters in ascending order such that $d_1 \le d_2 \le \dots \le d_N$. The continuous range of candidate heights $H$ to explore is then mathematically bounded by:
\begin{equation}
    H \in \left[ d_1 + \delta_{\min}, \;\; \sum_{i=1}^{N} d_i + \delta_{\max} \right]
\end{equation}
where $\delta_{\min} = -1.0$ and $\delta_{\max} = 1.0$ represent safety offset buffers.

\paragraph{Stage 1: Coarse Evaluation} 
The solver executes a wide sweep across the initialized bounds using a step size of $\Delta h_1 = 0.1$. To maintain high execution throughput during this exploratory phase, each candidate height point is allocated a modest optimization computation budget of $t_1 = 20$ seconds. Because \texttt{Sparrow} relies on stochastic meta-heuristics, the execution is repeated $n_1 = 3$ independent times per height coordinate, and the maximum area utilization score is recorded. We extract the height from this phase that maximized the occupancy ratio.

\paragraph{Stage 2: Refined Optimization} 
Once the primary operational region is identified, we perform a localized grid search centered directly around $h^*_{\text{stage1}}$ within a narrow window of $\pm \Delta h_1$. To achieve maximal placement precision, this phase utilizes a significantly finer step resolution of $\Delta h_2 = 0.03$. 

To handle the stochastic variance of the underlying placement heuristic and uncover peak performance configurations, the computation budget is expanded to $t_2 = 120$ seconds, and the engine executes $n_2 = 10$ independent stochastic runs per coordinate. Empirically, scaling the runtime budget beyond 120 seconds for instances containing up to 8 items yielded diminishing returns and failed to produce further area utilization improvements, confirming that this budget fully captures the baseline's optimal scaling curve. The absolute best layout discovered across all iterations in Stage 2 is selected as the final baseline solution.

\subsection{SVGNest Instantiation}
We apply the same orientation lock ($720$ bins, $0.5^\circ$) and the same diameter-derived bounds to \texttt{SVGNest}. Because the genetic engine is also a strip solver internally, the two-stage wrapper is extended from a 1D height search to a joint $(W,H)$ grid: Stage~1 cheaply scans candidate bins, and Stage~2 restarts the engine in the best Stage~1 bins. Layouts are scored by occupancy of the compacted axis-aligned bounding box of the placed parts, the same unconstrained metric as PoT, and this AABB occupancy is also used as the inner genetic fitness in place of the engine's native strip-packing fitness.

\section{Detailed Experimental Infrastructure and Results}
\label{app:experimental_details}

This section provides a granular overview of our experimental evaluation framework, detailing the underlying computing infrastructure, inference runtimes, and individual seed-by-seed area utilization logs across all hold-out benchmark instances.
\subsection{Hardware Configurations and Inference Runtimes}
To ensure a rigorous and reproducible performance comparison, we document the underlying computing infrastructure and absolute wall-clock execution budgets allocated for all three methods using strictly reproducible physical hardware specifications:

\begin{itemize}
    \item \textbf{Handcrafted Heuristic Baseline (\texttt{Sparrow}):} As a CPU-bound meta-heuristic optimization routine, the adapted multi-stage grid search for \texttt{Sparrow} was evaluated using a high-performance symmetric multiprocessing (SMP) server configuration powered by dual 4th Generation AMD EPYC 9004-series (Genoa) processors. This server provides a total of 180 physical computing cores and 360 concurrent hardware execution threads (via Symmetric Multithreading, SMT), backed by 1440 GB of high-speed DDR5 system memory. The total sequential processing budget consumed by \texttt{Sparrow} was approximately \textbf{1 hour on 364~CPUs} per execution seed for each evaluated tier.
    \item \textbf{Open-source Genetic Baseline (\texttt{SVGNest}):} We evaluate a scriptable Deepnest implementation of the \texttt{SVGNest} engine under the same two-stage container search and $0.5^\circ$ orientation grid, using AABB occupancy as the fitness criterion. Each seed ran for approximately \textbf{1 hour on 360~CPUs} per evaluated tier, matching the Sparrow compute regime.
    \item \textbf{Learned Policy Framework (Ours):} Neural network forward passes, continuous coordinate tokenizations, and spatial cross-attention maps for our PoT-driven agent were executed using an advanced heterogeneous compute node. GPU acceleration was provided by an array of 8 NVIDIA RTX 6000 Blackwell GPUs. These accelerators feature fifth-generation Tensor Cores and second-generation Transformer Engines supporting FP6 and FP4 precision, delivering high compute density and memory bandwidth. To coordinate the reinforcement learning framework, the node leverages AMD EPYC "Turin" CPUs providing 384 vCPUs and 1.4 TB of DDR5 memory, yielding a balanced ratio of 48 vCPUs per GPU. This hybrid hardware topology enables highly efficient asynchronous processing: the multi-core CPU architecture handles distributed orchestration, precomputed geometric features, and parallel environment simulations, while the GPU cluster executes the batched forward predictions of the model. The inference budget for PoT was approximately \textbf{10 hours on 8~GPUs} per execution seed for each evaluated tier.
\end{itemize}

Tables~\ref{tab:p4_all_seeds} and~\ref{tab:p8_all_seeds} document the exact area utilization rates achieved across 3 independent random execution seeds for PoT, \texttt{Sparrow}, and \texttt{SVGNest} over the 15 hold-out instances, together with the per-instance mean over those seeds. In each row, \textbf{bold} marks the highest of the three method means, and \underline{underline} marks the best occupancy among all nine seed runs.

\begin{table}[H]
\centering
\scriptsize
\caption{Granular area utilization (\%) on the 4-polygon tier (P4). Bold: highest mean over 3 seeds. Underline: best occupancy among all nine seed runs.}
\label{tab:p4_all_seeds}
\setlength{\tabcolsep}{2.0pt}
\renewcommand{\arraystretch}{0.95}
\resizebox{\textwidth}{!}{%
\begin{tabular}{lccccc c cccc c cccc}
\toprule
& \multicolumn{4}{c}{\textbf{Ours (PoT)}} & & \multicolumn{4}{c}{\textbf{Sparrow}} & & \multicolumn{4}{c}{\textbf{SVGNest}} \\
\cmidrule(lr){2-5} \cmidrule(lr){7-10} \cmidrule(lr){12-15}
\textbf{Instance ID} & \textbf{S1} & \textbf{S2} & \textbf{S3} & \textbf{Mean} & & \textbf{S1} & \textbf{S2} & \textbf{S3} & \textbf{Mean} & & \textbf{S1} & \textbf{S2} & \textbf{S3} & \textbf{Mean} \\
\midrule
Instance 00 & 78.08\% & \underline{78.10\%} & 78.07\% & \textbf{78.08\%} & & 77.81\% & 77.81\% & 77.81\% & 77.81\% & & 71.60\% & 73.29\% & 71.79\% & 72.23\% \\
Instance 01 & 78.14\% & \underline{78.16\%} & \underline{78.16\%} & \textbf{78.15\%} & & 78.00\% & 78.01\% & 78.01\% & 78.01\% & & 67.77\% & 68.64\% & 70.38\% & 68.93\% \\
Instance 02 & 82.01\% & 82.02\% & \underline{82.03\%} & \textbf{82.02\%} & & 81.87\% & 81.88\% & 81.89\% & 81.88\% & & 74.48\% & 73.80\% & 74.68\% & 74.32\% \\
Instance 03 & 77.67\% & \underline{77.76\%} & 77.68\% & 77.70\% & & 77.70\% & 77.74\% & 77.71\% & \textbf{77.72\%} & & 68.30\% & 67.59\% & 66.88\% & 67.59\% \\
Instance 04 & \underline{82.23\%} & 82.20\% & 82.18\% & \textbf{82.20\%} & & 81.81\% & 81.90\% & 81.80\% & 81.84\% & & 72.32\% & 72.47\% & 71.18\% & 71.99\% \\
Instance 05 & 84.26\% & 84.27\% & 84.26\% & 84.26\% & & \underline{84.29\%} & 84.28\% & 84.27\% & \textbf{84.28\%} & & 78.06\% & 77.13\% & 78.73\% & 77.97\% \\
Instance 06 & 75.20\% & 75.34\% & \underline{75.62\%} & \textbf{75.39\%} & & 75.15\% & 75.14\% & 75.15\% & 75.15\% & & 65.81\% & 67.43\% & 66.25\% & 66.49\% \\
Instance 07 & 79.59\% & 79.58\% & 79.84\% & 79.67\% & & 79.84\% & \underline{79.87\%} & 79.85\% & \textbf{79.85\%} & & 69.54\% & 68.86\% & 72.69\% & 70.36\% \\
Instance 08 & 80.96\% & 80.94\% & \underline{81.04\%} & \textbf{80.98\%} & & 80.94\% & 80.90\% & 80.94\% & 80.93\% & & 72.03\% & 74.45\% & 73.55\% & 73.34\% \\
Instance 09 & 80.43\% & \underline{80.58\%} & 80.37\% & 80.46\% & & 80.56\% & 80.51\% & 80.57\% & \textbf{80.55\%} & & 71.93\% & 70.72\% & 70.36\% & 71.00\% \\
Instance 10 & 82.95\% & 82.92\% & \underline{83.05\%} & \textbf{82.97\%} & & 82.92\% & 82.92\% & 82.96\% & 82.93\% & & 74.11\% & 72.96\% & 73.80\% & 73.63\% \\
Instance 11 & 77.48\% & 77.37\% & 77.50\% & 77.45\% & & \underline{77.75\%} & \underline{77.75\%} & 77.73\% & \textbf{77.74\%} & & 71.00\% & 67.98\% & 69.61\% & 69.53\% \\
Instance 12 & 81.07\% & \underline{81.12\%} & 81.06\% & \textbf{81.08\%} & & 80.46\% & 80.48\% & 80.47\% & 80.47\% & & 70.71\% & 70.92\% & 69.78\% & 70.47\% \\
Instance 13 & 93.02\% & 93.02\% & 93.03\% & \textbf{93.02\%} & & 92.59\% & \underline{93.10\%} & 92.63\% & 92.77\% & & 85.34\% & 83.51\% & 84.67\% & 84.51\% \\
Instance 14 & 84.20\% & \underline{84.22\%} & 84.20\% & \textbf{84.21\%} & & 84.09\% & 84.07\% & 84.10\% & 84.09\% & & 72.92\% & 72.33\% & 70.20\% & 71.81\% \\
\bottomrule
\end{tabular}%
}
\end{table}

\begin{table}[H]
\centering
\scriptsize
\caption{Granular area utilization (\%) on the 8-polygon tier (P8). Bold: highest mean over 3 seeds. Underline: best occupancy among all nine seed runs.}
\label{tab:p8_all_seeds}
\setlength{\tabcolsep}{2.0pt}
\renewcommand{\arraystretch}{0.95}
\resizebox{\textwidth}{!}{%
\begin{tabular}{lccccc c cccc c cccc}
\toprule
& \multicolumn{4}{c}{\textbf{Ours (PoT)}} & & \multicolumn{4}{c}{\textbf{Sparrow}} & & \multicolumn{4}{c}{\textbf{SVGNest}} \\
\cmidrule(lr){2-5} \cmidrule(lr){7-10} \cmidrule(lr){12-15}
\textbf{Instance ID} & \textbf{S1} & \textbf{S2} & \textbf{S3} & \textbf{Mean} & & \textbf{S1} & \textbf{S2} & \textbf{S3} & \textbf{Mean} & & \textbf{S1} & \textbf{S2} & \textbf{S3} & \textbf{Mean} \\
\midrule
Instance 00 & 83.39\% & 82.44\% & 82.45\% & 82.76\% & & \underline{86.87\%} & 86.20\% & 86.22\% & \textbf{86.43\%} & & 71.24\% & 70.19\% & 70.19\% & 70.54\% \\
Instance 01 & \underline{83.95\%} & 83.80\% & 83.52\% & \textbf{83.76\%} & & 83.02\% & 82.59\% & 83.87\% & 83.16\% & & 71.69\% & 73.74\% & 71.16\% & 72.20\% \\
Instance 02 & 82.39\% & 82.36\% & 81.58\% & 82.11\% & & \underline{85.57\%} & 84.91\% & 85.19\% & \textbf{85.22\%} & & 74.52\% & 74.50\% & 72.44\% & 73.82\% \\
Instance 03 & 83.73\% & 84.75\% & 83.92\% & 84.13\% & & 85.71\% & 85.57\% & \underline{85.92\%} & \textbf{85.73\%} & & 65.47\% & 68.42\% & 67.74\% & 67.21\% \\
Instance 04 & 84.16\% & 84.93\% & \underline{86.52\%} & \textbf{85.20\%} & & 83.56\% & 83.75\% & 84.27\% & 83.86\% & & 72.17\% & 72.38\% & 71.68\% & 72.08\% \\
Instance 05 & 82.89\% & 84.21\% & 83.45\% & 83.52\% & & \underline{85.87\%} & 85.18\% & 85.28\% & \textbf{85.44\%} & & 73.44\% & 69.91\% & 74.91\% & 72.75\% \\
Instance 06 & 84.74\% & 84.00\% & 85.37\% & 84.70\% & & \underline{88.10\%} & 87.98\% & 87.37\% & \textbf{87.82\%} & & 72.29\% & 71.84\% & 72.37\% & 72.17\% \\
Instance 07 & 80.92\% & 81.68\% & 81.38\% & 81.33\% & & \underline{86.01\%} & 85.24\% & 85.16\% & \textbf{85.47\%} & & 67.11\% & 66.85\% & 65.75\% & 66.57\% \\
Instance 08 & 81.46\% & 82.64\% & 82.78\% & 82.29\% & & 84.72\% & 83.99\% & \underline{85.03\%} & \textbf{84.58\%} & & 67.59\% & 64.61\% & 67.00\% & 66.40\% \\
Instance 09 & 74.75\% & 74.25\% & 74.58\% & 74.53\% & & 84.29\% & 83.92\% & \underline{84.63\%} & \textbf{84.28\%} & & 59.73\% & 60.56\% & 58.86\% & 59.72\% \\
Instance 10 & 80.58\% & 81.32\% & 80.85\% & 80.92\% & & 83.62\% & 84.67\% & \underline{84.69\%} & \textbf{84.33\%} & & 69.10\% & 66.07\% & 67.14\% & 67.44\% \\
Instance 11 & 83.32\% & 83.06\% & 83.89\% & 83.42\% & & \underline{86.15\%} & 84.96\% & 85.49\% & \textbf{85.53\%} & & 71.46\% & 70.22\% & 71.36\% & 71.01\% \\
Instance 12 & 77.82\% & 78.25\% & 77.22\% & 77.76\% & & \underline{84.96\%} & 84.76\% & 84.28\% & \textbf{84.67\%} & & 64.36\% & 62.51\% & 63.43\% & 63.43\% \\
Instance 13 & 81.23\% & 81.03\% & 80.56\% & 80.94\% & & 84.63\% & 84.19\% & \underline{84.73\%} & \textbf{84.52\%} & & 67.11\% & 69.77\% & 66.03\% & 67.64\% \\
Instance 14 & 78.32\% & 79.90\% & 80.08\% & 79.43\% & & \underline{88.08\%} & 86.27\% & 86.18\% & \textbf{86.84\%} & & 68.02\% & 64.84\% & 64.99\% & 65.95\% \\
\bottomrule
\end{tabular}%
}
\end{table}


\subsection{Inference Strategy}

We implemented the inference strategy by keeping a buffer of the best $0.1\%$ of configurations and sampling from it $30\%$ of the time to set the episode's initial configuration. 

We show in the following table how many of the 15 instances' best configurations from both P4 and P8 were found using this buffer.

\begin{table}[htbp]
    \centering
    \begin{tabular}{lcc}
        \toprule
        \textbf{Initialization Source} & \textbf{P4 (Mean)} & \textbf{P8 (Mean)} \\
        \midrule
        Random Initialization & 2.00 & 2.33 \\
        Buffer Resampling & 13.00 & 12.67 \\
        \bottomrule
    \end{tabular}
    \caption{Distribution of the top 15 highest-performing solutions discovered across the P4 and P8 instance sets, categorized by their episode initialization source.}
    \label{tab:buffer_strategy}
\end{table}

\end{document}